%% file: camera_ready.tex
\def\year{2021}\relax
\documentclass[letterpaper]{article} 
\usepackage{arxiv}  
\usepackage{times}  
\usepackage{helvet} 
\usepackage{courier}  
\usepackage[hyphens]{url}  
\usepackage{graphicx} 
\usepackage[switch]{lineno}
\usepackage{amsfonts}
\input{config}

\usepackage{natbib}  
\usepackage{caption} 
\usepackage{hyperref}
\hypersetup{colorlinks=true,citecolor=blue}

\title{Learning Energy-Based Model \\ with Variational Auto-Encoder as Amortized Sampler}
\author{Jianwen Xie, Zilong Zheng, Ping Li\\}
\affiliations{Cognitive Computing Lab\\ 
Baidu Research\\
10900 NE 8th St. Bellevue, WA 98004, USA\\
\{jianwen.kenny, zlzheng.cs, pingli98\}@gmail.com
}

\begin{document}

\maketitle

\begin{abstract}
Due to the intractable partition function, training energy-based models (EBMs) by maximum likelihood requires Markov chain Monte Carlo (MCMC) sampling to approximate the gradient of the Kullback-Leibler divergence between data and model distributions. However, it is non-trivial to sample from an EBM because of the difficulty of mixing between modes. In this paper, we propose to learn a variational auto-encoder (VAE) to initialize the finite-step MCMC, such as Langevin dynamics that is derived from the energy function, for efficient amortized sampling of the EBM. With these amortized MCMC samples, the EBM can be trained by maximum likelihood, which follows an ``analysis by synthesis'' scheme; while the VAE learns from these MCMC samples via variational Bayes. We call this joint training algorithm the \text{variational MCMC teaching}, in which the VAE chases the EBM toward data distribution. We interpret the learning algorithm as a dynamic alternating projection in the context of information geometry. Our proposed models can generate samples comparable to GANs and EBMs. Additionally, we demonstrate that our model can learn effective probabilistic distribution toward supervised conditional learning tasks. 

\end{abstract}

\section{Introduction}
Generative modeling of high-dimensional data is a very challenging and fundamental problem in both computer vision and machine learning communities. 
Energy-based generative model~\cite{zhu1998filters,Lecun2006} with the energy function parameterized by a deep neural network 
was first proposed by~\citet{XieLuICML}, and has been drawing attention in the recent literature~\cite{XieCVPR17,gao2018learning, xie2018learning,xie2019learning, du2019implicit,nijkamp2019learning,grathwohl2019your}, not only for its empirically powerful ability to learn highly complex probability distribution, but also for its theoretically fascinating aspects of representing high-dimensional data. Successful applications with energy-based generative frameworks have been witnessed in the field of computer vision, for example, video synthesis \cite{XieCVPR17,xie2019learning}, 3D volumetric shape synthesis \cite{xie2018learning,Xie2020GenerativeVL}, unordered point cloud synthesis \cite{xie2021GPointNet}, supervised image-to-image translation \cite{xie2019cooperative}, and unpaired cross-domain visual translation \cite{xie2021cycleCoopNets}. Other applications can be seen in natural language processing \cite{bakhtin2021residual}, biology~\cite{ingraham2018learning, du2019energy}, and inverse optimal control~\cite{xu2019energy}. 

Energy-based generative models directly define an unnormalized probability density that is an exponential of the negative energy function, where the energy function maps the input variable to an energy scalar. 
Training  an energy-based model (EBM) from observed data corresponds to finding an energy function, where observed data are assigned lower energies than unobserved ones. 
Synthesizing new data from the energy-based probability density can be achieved by a gradient-based Markov chain Monte Carlo (MCMC) method, which is an implicit and iterative generation process, to find low energy regions of the learned energy landscape; we refer readers to two excellent textbooks~\cite{liu2008monte,barbu2020monte} and numerous references therein. Energy-based generative models, therefore, unify the generation and learning processes in a single~model. 

A persisting challenge in training an EBM of high-dimensional data via maximum likelihood estimation (MLE) is the calculation of the normalizing constant or the partition function, which requires a computationally intractable integral. Therefore, an MCMC sampling procedure, such as the Langevin dynamics or Hamiltonian Monte Carlo~\cite{neal2011mcmc}, from the EBMs is typically used to approximate the gradient of the partition function during the model training. However, the MCMC is computationally expensive or even impractical, especially if the target distribution has multiple modes separated by highly low probability regions. In such a case, traversing modes becomes very difficult and unlikely because different MCMC chains easily get trapped by different local modes.

To tackle the above challenge, with the inspiration of the idea of amortized generation in~\citet{Bengio2016,xie2018cooperative}, we propose to train a directed latent variable model as an approximate sampler that generates samples by deterministic transformation of independent and identically distributed random samples drawn from Gaussian distribution. Such an ancestral sampler can efficiently provide a good initial point of the iterative MCMC sampling of the EBM to avoid a long computation time to generate convergent samples. We call this process of first running an ancestral sampling by a latent variable model and then revising the samples by a finite-step Langevin dynamics derived from an EBM the \textit{Ancestral Langevin Sampling} (ALS). ALS takes advantages of both Langevin sampling and ancestral sampling. First, because the ancestral sampler connects the low-dimensional Gaussian distribution with the high-dimensional data distribution, traversing modes of the data distribution becomes more tractable and practical by sampling from the low-dimensional latent space. Secondly, the Langevin sampler is an attractor-like dynamics that can refine the initial samples by attracting them to the local modes of the energy function, thus making the initially generated samples stabler and more likely configurations.

From the learning perspective, by comparing the difference between the observed examples and the ALS examples, the EBM can find its way to shift its density toward the data distribution via MLE. The ALS with a small number of Langevin steps can accelerate the training of the EBM in terms of convergence speed. To approximate the Langevin sampler and serves as a good MCMC initializer, the latent variable model learns from the evolving EBM by treating the ALS examples at each iteration as training data. Different from~\citet{Bengio2016,xie2018cooperative}, we follow the variational Bayes \cite{kingma2013auto} to train the latent variable model by recruiting an approximate but computationally efficient inference model, which is typically an encoder network. Specifically, after the EBM revises the initial examples provided by the latent variable model, the inference model infers the latent variables of the revised examples, and then the latent variable model updates its mapping function by regressing the revised examples on their corresponding inferred latent codes. The inference model and the latent variable model form a modified \ac{VAE}~\cite{kingma2013auto} that learns from evolving ALS samples, which are MCMC samples from the EBMs. In this framework, the EBM provides infinite batches of fresh MCMC examples as training data to the VAE model. The learning of the VAE are affected by the EBM. While providing help to the EBM in sampling, the VAE learns to chase the EBM, which runs towards the data distribution with the efficient sampling, ALS. Within the VAE, the inference model and the posterior of the latent variable model get close to each other via maximizing the variational lower bound of the log likelihood of the ALS samples. In other words, the latent variable model is trained with both variational inference of the inference model and MCMC teaching of the EBM. We call this the \textit{Variational MCMC teaching}. 

Moreover, the generative framework can be easily generalized to the conditional model by involving a conditional EBM and a conditional VAE, for representing a distribution of structured output given another structured input. This conditional model is very useful and can be applied to plenty of computer vision tasks, such as image inpainting etc.



\vspace{0.1in}

Concretely, our contributions can be summarized below:
\begin{enumerate}
\item We present a new framework to train \acf{EBMs}, where a VAE is jointly trained via MCMC teaching to fast initialize the Langevin dynamics of the EBM for its maximum likelihood learning. The amortized sampler is called \textit{ancestral Langevin sampler}.
\item Our model provides a new strategy that we call \textit{variational MCMC teaching} to train latent variable model, where an EBM and an inference model are simultaneously trained to provide infinite training examples and efficient approximate inference for the latent variable model, respectively. 
\item We naturally unify the maximum likelihood learning, variational inference, and MCMC teaching in a single framework to induce maximum likelihood learning of all the probability models. 
\item We provide an information geometric understanding of the proposed joint training algorithm. It can be interpreted as a dynamic alternating projection.   
\item We provide strong empirical results on unconditional image modeling and conditional predictive learning to corroborate the proposed method.
\end{enumerate}



\section{Related Work}

There are three types of interactions inside our model. The inference model and the latent variable model are trained in a variational inference scheme~\cite{kingma2013auto}, the \ac{EBM} and the latent variable model are trained in a cooperative learning scheme~\cite{xie2018cooperative}, and also the \ac{EBM} and the data distribution forms an MCMC-based maximum likelihood estimation or ``analysis by synthesis'' learning scheme~\cite{XieLuICML,du2019implicit,nijkamp2019learning}. 

\vspace{0.05in}
\textbf{Energy-based density estimation.} The maximum likelihood estimation of the energy-based model~\cite{zhu1998filters,wu2000texture,Lecun2006,hinton2012practical,xie2014learning,LuZhuWu2016,XieLuICML},
 follows what~\citet{grenander2007pattern} call ``analysis by synthesis'' scheme, where, at each iteration, the computation of the gradient of the log-likelihood requires MCMC sampling, such as the Gibbs sampling~\cite{geman1984stochastic}, or Langevin dynamics. To overcome the computational hurdle of MCMC, the contrastive divergence~\cite{Hinton2002a}, which is an approximate maximum-likelihood, initializes the MCMC with training data in learning the EBM. The noise-contrastive estimation~\cite{gutmann2012noise} of the \ac{EBM} turns a generative learning problem into a discriminative learning one by preforming nonlinear logistic regression to discriminate the observed examples from some artificially generated noise examples.~\citet{nijkamp2019learning} propose to learn \ac{EBM} with non-convergent non-persistent short-run MCMC as a flow-based generator, which can be useful for synthesis and reconstruction. The training of the \ac{EBM} in our framework still follows ``analysis by synthesis'', except that the synthesis is performed by the \textit{ancestral Langevin sampling}.

\vspace{0.05in}
\textbf{Training an \ac{EBM} jointly with a complementary model.}  To avoid MCMC sampling of the \ac{EBM},~\citet{Bengio2016} approximate it by a latent variable model trained by minimizing the Kullback-Leibler  (KL) divergence from the latent variable model to the EBM. It involves an intractable entropy term, which is problematic if it is ignored.
The gap between the latent variable model and the EBM due to their imbalanced model design may still cause bias or model collapse in training. 
We bridge the gap by taking back the MCMC to serve as an attractor-like dynamics to refine any imperfection of the latent variable model in the learned VAE. 
\citet{xie2018cooperative, song2018learning} study a similar problem. In comparison with~\citet{xie2018cooperative}, which either uses another MCMC to compute the intractable posterior of the latent variable model or directly ignores the inference step for approximation in their experiments, our framework learns a tractable variational inference model for training the latent variable model. The proposed framework is a variant of cooperative networks in ~\citet{xie2018cooperative}.

\section{Preliminary}
In this section, we present the backgrounds of \acf{EBMs} and \acf{VAEs}, which will serve as foundations of the proposed framework. 
\subsection{\ac{EBM} and Analysis by Synthesis}
\label{sec:ebm_def}
Let $x \in \mathbb{R}^{D}$ be the high-dimensional random variable, such as an input image. An EBM (also called Markov random field, Gibbs distribution, or exponential family model), with an energy function $U_{\theta}(x)$ and a set of trainable parameters $\theta$, learns to associate a scalar energy value to each configuration of the random variable, such that more plausible configurations (e.g., observed training images) are assigned lower energy values. Formally, an EBM is defined as a probability density with the following form:
\begin{eqnarray}
p_{\theta}(x) = \frac{1}{Z(\theta)} \exp[-U_{\theta}(x)], 
\label{eq:ebm}
\end{eqnarray}  
where $Z(\theta)= \int \exp[-U_{\theta}(x)]dx$ is a normalizing constant or a partition function depending on $\theta$, and is analytically intractable to calculate due to high dimensionality of $x$. Following the EBM introduced in~\citet{XieLuICML}, we can parameterize $U_{\theta}(x)$ by a bottom-up  ConvNet with trainable weights $\theta$ and scalar output. 

Assume a training dataset $\mathcal{D}=\{x_i, i=1,...,n\}$ is given and each data point is sampled from an unknown distribution $p_{\rm data}(x)$. In order to use the \ac{EBM} $p_{\theta}(x)$ to estimate the data distribution $p_{\rm data}(x)$, we can minimize the negative log-likelihood of the observed data $L(\theta, \mathcal{D})=-\frac{1}{n} \sum_{i=1}^{n} \log p_{\theta}(x_i)$, or equivalently the KL-divergence between the two distributions $\KL(p_{\rm data}(x)||p_{\theta}(x))$ by gradient-based optimization methods. The gradient to update parameters $\theta$ is computed by the following formula  
\begin{eqnarray}
\begin{aligned}
& \frac{\partial}{\partial \theta}\KL(p_{\rm data}(x)||p_{\theta}(x))\\
 =& \E_{x \sim p_{\rm data}(x)} \left[ \frac{\partial U_{\theta}(x)}{\partial \theta} \right] - \E_{\tilde{x} \sim p_{\theta}(x)} \left[ \frac{\partial U_{\theta}(\tilde{x})}{\partial \theta} \right].
\label{eq:gradient}
\end{aligned}
\end{eqnarray}

The two expectations in Eq. (\ref{eq:gradient}) are approximated by averaging over the observed examples $\{x_i\}$ and the synthesized examples $\{\tilde{x}_i\}$ that are sampled from the model $p_{\theta}(x)$, respectively. This will lead to an “analysis by synthesis” algorithm that
iterates a synthesis step for image sampling and an analysis step for parameter learning. 

Drawing samples from \ac{EBM}s typically requires Markov chain Monte Carlo (MCMC) methods. If the data distribution $p_{\rm data}(x)$ is complex and multimodal, the MCMC sampling from the learned model is challenging because it may take a long time to mix between modes. Thus, the ability to generate efficient and fair examples from the model becomes the key to training successful \ac{EBMs}. In this paper, we will study amortized sampling for efficient training of the EBMs. 

\subsection{Latent Variable Model and Variational Inference}
Consider a directed latent variable model of the form 
\begin{eqnarray}
&& z \sim \mathcal{N}(0,I_d), x=g_{\alpha}(z)+ \epsilon, \epsilon \sim \mathcal{N}(0,\sigma^2 I_D), \label{eq:lvm}
\end{eqnarray}
where $z \in \mathbb{R}^d$ is a $d$-dimensional vector of latent variables following a Gaussian distribution $\mathcal{N}(0,I_d)$, $I_d$ is a $d$-dimensional identity matrix, $g_{\alpha}$ is a nonlinear mapping function that is parameterized by a top-down deep neural network with trainable parameters $\alpha$, and $\epsilon \in \mathbb{R}^D$ is the residual noise that is independent of $z$. 

The marginal distribution of the model in Eq.~(\ref{eq:lvm}) is $q_{\alpha}(x)=\int q_{\alpha}(x|z)q(z) dz$, where the prior distribution $q(z)=\mathcal{N}(0,I_d)$ and the conditional distribution of $x$ given $z$ is $q_{\alpha}(x|z)=\mathcal{N}(g_{\alpha}(z), \sigma^2I_{D})$. The posterior distribution is $q_{\alpha}(z|x)=q_{\alpha}(z,x)/q_{\alpha}(x)=q_{\alpha}(x|z)q(z)/q_{\alpha}(x)$. Both posterior distribution $q_{\alpha}(z|x)$ and marginal distribution $q_{\alpha}(x)$ are analytically intractable. As in~\citet{han2017alternating}, the model can be learned by maximum likelihood estimation or equivalently minimizing the KL-divergence $\KL(p_{\rm data}(x)||q_{\alpha}(x))$, whose gradient is given by
\begin{eqnarray}
\begin{aligned}
&\frac{\partial}{\partial \alpha}\KL(p_{\rm data}(x)||q_{\alpha}(x)) \\
=& \E_{p_{\rm data}(x)q_{\alpha}(z|x)} \left[ -\frac{\partial}{\partial \alpha} \log q_{\alpha}(z,x) \right]. 
\label{eq:generator_gradient}
\end{aligned}
\end{eqnarray}

MCMC methods can be used to compute the gradient in Eq.~(\ref{eq:generator_gradient}). For each data point $x_i$ sampled from the data distribution, we infer the corresponding latent variable $z_i$ by drawing samples from $q_{\alpha}(z|x)$ via MCMC methods, then the expectation term can be approximated by averaging over the sampled pairs $\{x_i,z_i\}$.     However, MCMC sampling of the posterior distribution may also take a long time to converge. To avoid MCMC sampling from $q_{\alpha}(z|x)$, \ac{VAE} \cite{kingma2013auto} approximates $q_{\alpha}(z|x)$ by a tractable inference network, for example, a
multivariate Gaussian with a diagonal covariance structure
 $\pi_{\beta}(z|x) \sim \mathcal{N}(\mu_{\beta}(x), {\rm diag}(v_{\beta}(x)))$, where both $\mu_{\beta}(x)$ and $v_{\beta}(x)$ are $d$-dimensional outputs of encoding bottom-up networks of data point $x$, with trainable parameters $\beta$. With this reparameterization trick, the objective of VAE becomes to find $\alpha$ and $\beta$ to minimize
\begin{eqnarray}
\begin{aligned}
&\KL(p_{\rm data}(x)\pi_{\beta}(z|x)||q_{\alpha}(z,x)) \\
= &\KL(p_{\rm data}(x)||q_{\alpha}(x)) + \KL(\pi_{\beta}(z|x)||q_{\alpha}(z|x)),
\label{eq:vae}
\end{aligned}
\end{eqnarray}   
which is a modification of the maximum likelihood estimation objective. Minimizing the left-hand side in Eq. (\ref{eq:vae}) will also lead to a minimization of the first KL-divergence on the right-hand side, which is the maximum likelihood estimation objective in Eq. (\ref{eq:generator_gradient}). In this paper, we will propose to learn a latent variable model in the context of VAE as amortized sampler to train the EBM. 

\section{Methodology} 
\label{sec:model} 
We study to learn an EBM via MLE with a VAE as amortized sampler. The amortized sampler is achieved by integrating the latent variable model (the generator network in VAE) and the short-run MCMC of the EBM. We propose to jointly train EBM and VAE via  \textit{variational MCMC~teaching}. 

\subsection{Ancestral Langevin Sampling}
To learn the energy-based generative model in Eq.~(\ref{eq:ebm}) and compute the gradient in Eq.~(\ref{eq:gradient}), 
we might bring in a directed latent variable model $q_{\alpha}(x)$ to serve as a fast non-iterative sampler to initialize the iterative MCMC sampler guided by the energy function $U_{\theta}$, for the sake of efficient MCMC convergence and mode traversal of the EBM. In our paper, we call the resulting amortized sampler the \textit{ancestral Langevin sampler}, which draws a sample by first (i) sampling an initial example $\hat{x}$ via ancestral sampling, and then (ii) revising $\hat{x}$ with a finite-step Langevin update, that is 
\begin{eqnarray}
\begin{aligned} 
&{\rm (i)}~\hat{x}= g_{\alpha}(\hat{z}), ~\hat{z} \sim \mathcal{N}(0, I_{d}), \\
 &{\rm (ii)}~\tilde{x}_{t+1} = \tilde{x}_{t} - \frac{\delta^2}{2} \frac{\partial U_{\theta}(\tilde{x}_{t})}{\partial \tilde{x}} + \delta \mathcal{N}(0,I_D),   ~\tilde{x}_{0}=\hat{x}, 
\end{aligned} 
\label{eq:Langevin}
\end{eqnarray}
where $\hat{x}$ is the initial example generated by ancestral sampling, $\tilde{x}$ is the example generated by the Langevin dynamics, $t$ indexes the Langevin time step, and $\delta$ is the step size. The Langevin dynamics is equivalent to a stochastic gradient descent algorithm that seeks to find the minimum of the objective function defined by $U_{\theta}(x)$. 

Generally, in the original ``analysis by synthesis'' algorithm, the Langevin dynamics shown in Eq. (\ref{eq:Langevin})(ii) is initialized with a noise distribution, such as Gaussian distribution, i.e., $\tilde{x}_{0} \sim \mathcal{N}(0,I_D)$, and this usually takes a long time to converge and is also non-stable in practise because the gradient-based MCMC chains can get trapped in the local modes when exploring the model distribution. 

As to the \textit{ancestral Langevin sampling} in Eq. (\ref{eq:Langevin}), intuitively, if the latent variable model in Eq. (\ref{eq:Langevin})(i) can memorize the majority of the modes in $p_{\theta}(x)$ by low dimensional codes $\hat{z}$, then we can easily traverse among modes of the model distribution by simply sampling from $p(\hat{z})=\mathcal{N}(0,I_d)$, because $p(\hat{z})$ is much smoother than $p_{\theta}(x)$. The short-run Langevin dynamics initialized with the output $\hat{x}$ of the latent variable model emphasizes on refining the detail of $\hat{x}$ by further searching for a better mode $\tilde{x}$ around $\hat{x}$. Ideally, if $p_{\theta}(x)$ and $q_{\alpha}(x)$ fit the data distribution $p_{\text{data}}(x)$ perfectly, the example $\hat{x}$ produced by the ancestral sampling will be exactly on the modes of $U_{\theta}(x)$. In this case, the following Langevin revision will not change the $\hat{x}$, i.e., $\tilde{x}=\hat{x}$. Otherwise, the Langevin update will further improve $\hat{x}$. 


\subsection{Variational MCMC Teaching}
With $\{\tilde{x}_i\}_{i=1}^{\tilde{n}} \sim p_{\theta}(x)$ via \textit{ancestral Langevin sampling} in Eq.~(\ref{eq:Langevin}), we can compute the gradient in Eq.~(\ref{eq:gradient}) by
\begin{eqnarray}
\begin{aligned} 
&\frac{\partial}{\partial \theta}\KL(p_{\rm data}(x)||p_{\theta}(x)) \\
\approx & \frac{1}{n} \sum_{i=1}^{n} \frac{\partial U_{\theta}(x_i)}{\partial \theta} - \frac{1}{\tilde{n}} \sum_{i=1}^{\tilde{n}} \frac{\partial U_{\theta}(\tilde{x}_i)}{\partial \theta}
 \label{eq:gradient2}
\end{aligned} 
\end{eqnarray}

\noindent and then update $\theta$ by Adam~\cite{kingma2014adam}. Consider in 
this iterative algorithm, the current model parameter $\theta$ and $\alpha$ are $\theta_t$ and $\alpha_t$ respectively. We use $\M_{\theta_{t}}$ to denote the Markov transition kernel of a finite-step Langevin dynamics that samples from the current distribution $p_{\theta_{t}}(x)$. We also use $\M_{\theta_{t}}q_{\alpha_{t}}(x)=\int M_{\theta_{t}}(x',x)q_{\alpha_{t}}(x')dx'$ to denote the marginal distribution obtained by running $\M_{\theta_{t}}$ initialized from current $q_{\alpha_{t}}(x)$. The MCMC-based MLE training of $\theta$ seeks to minimize the following objective at each iteration
\begin{eqnarray} 
\begin{aligned} 
\theta_{t+1} &= \argmin_{\theta} [ \KL(p_{\rm data}(x){\parallel}p_{\theta}(x))\\ 
&- \KL(\M_{\theta_{t}} q_{\alpha_{t}}(x){\parallel} p_{\theta}(x)) ], 
\label{eq:objective_mle}
\end{aligned} 
\end{eqnarray} 
which is considered as a modified contrastive divergence in \citet{xie2018cooperative, xie2016cooperative}.    
Meanwhile, $q_{\alpha_{t+1}}(x)$ is learned based on how the finite steps of Langevin $\M_{\theta_t}$ revises the initial example $\{\hat{x}_i\}$ generated by $q_{\alpha_{t}}(x)$ to mimic the Langevin sampling. This is the energy-based MCMC teaching~\cite{xie2018cooperative, xie2016cooperative} of $q_{\alpha}(x)$ .

Although $q_{\alpha}(x)$ initializes the Langevin sampling of $\{\tilde{x}_i\}$, the corresponding latent variables of $\{\tilde{x}_i\}$ are no longer $\{\hat{z}_i\}$. 
To retrieve the latent variables of $\{\tilde{x}_i\}$, we propose to infer $\tilde{z} \sim \pi_{\beta}(z|\tilde{x})$, which is an approximate tractable inference network, and then learn $\alpha$ from complete data $\{\tilde{z}_i, \tilde{x}_i\}_{i=1}^{\tilde{n}}$ to minimize $\sum_i ||\tilde{x}_i-g_{\alpha}(\tilde{z}_i)||^{2}$ (or equivalently maximize $\sum_i \log q_{\alpha}(\tilde{z}_i,\tilde{x}_i)$). To ensure $\pi_{\beta}(z|\tilde{x})$ to be an effective inference network that mimics the computation of the true inference procedure $\tilde{z} \sim q_{\alpha}(z|\tilde{x})$, we simultaneously learn $\beta$ by minimizing $ \KL(\pi_{\beta}(z|x)||q_{\alpha}(z|x))$, i.e., the reparameterization trick of the variational inference~of~$q_{\alpha}(x)$.

The learning of $\pi_{\beta}(z|x)$ and $q_{\alpha}(x|z) $ forms a VAE that treats $\{\tilde{x}_i\}$ as training examples. Because $\{\tilde{x}_i\}$ are dependent on $\theta$ and vary during training, the objective function of the VAE is non-static. This is essentially different from the original VAE that has a fixed training data. Suppose we have $\{\tilde{x}_i\}_{i=1}^{\tilde{n}} \sim \M_{\theta_{t}}q_{\alpha_t}(x)$ at the current iteration $t$, the VAE objective in our framework is the minimization of variational lower bound of the negative log likelihood of $\{\tilde{x}_i\}_{i=1}^{\tilde{n}}$, i.e.,
\begin{eqnarray} 
\begin{aligned} 
L(\alpha, \beta) &= \sum_{i=1}^{\tilde{n}} [ -\log q_{\alpha}(\tilde{x}_i) \\ 
&+ \gamma \KL(\pi_{\beta}(z_i|\tilde{x}_i)||q_{\alpha}(z_i|\tilde{x}_i))],
\label{eq:vae_lower_bound}
\end{aligned} 
\end{eqnarray} 
where $\gamma$ is a hyper-parameter that specifies the importance of the KL-divergence term. Since when $\tilde{n} \rightarrow \infty$, we have
\begin{eqnarray}\notag
\min_{\alpha} \sum_{i=1}^{\tilde{n}} [-\log q_{\alpha}(\tilde{x}_i)] = \min_{\alpha} \KL(\M_{\theta_{t}}q_{\alpha_t}(x)||q_{\alpha}(x)), 
\end{eqnarray}
thus Eq.~(\ref{eq:vae_lower_bound}) is equivalent to minimizing 
\begin{eqnarray}
\begin{aligned}
&\KL(\M_{\theta_{t}}q_{\alpha_t}(x)||q_{\alpha}(x)) + \KL(\pi_{\beta}(z|x)||q_{\alpha}(z|x)) \\
= &\KL(\M_{\theta_{t}}q_{\alpha_t}(x)\pi_{\beta}(z|x)||q_{\alpha}(x|z)q(z)) \label{eq:vae_lower_bound2}. 
\end{aligned}
\end{eqnarray}  
Unlike the objective function of the maximum likelihood estimation $\KL(\M_{\theta_{t}}q_{\alpha_t}(x)||q_{\alpha}(x))$, which involves intractable marginal distribution $q_{\alpha}(x)$, the variational objective function is the KL-divergence between the joint distributions, which is tractable because $\pi_{\beta}(z|x)$ parameterized by an encoder is tractable. In comparison with the original VAE objective in Eq.~(\ref{eq:vae}), our VAE objective in  Eq.~(\ref{eq:vae_lower_bound2}) replaces $p_{\rm data}(x)$ by $\M_{\theta_{t}}q_{\alpha_t}(x)$. 
At each iteration, minimizing the variational objective in Eq. (\ref{eq:vae_lower_bound2}) will eventually decrease $\KL(\M_{\theta_{t}}q_{\alpha_t}(x)||q_{\alpha}(x))$. 
Since $q_{\alpha}(x)$ is learned in the context of both MCMC teaching \cite{xie2016cooperative} and variational inference \cite{kingma2013auto}. We call this the \textit{variational MCMC teaching}.  Algorithm~\ref{alg:1} describes the proposed joint training algorithm of EBM and VAE. 

\begin{algorithm}[h!]
\caption{Cooperative training of EBM and VAE via variational MCMC teaching}
\label{alg:1}
\begin{algorithmic}[1]

\Require (a) training images $\{x_i\}_{i=1}^{n}$, (b) number of Langevin steps $l$ 
\Ensure (a) model parameters $\{\theta,\alpha, \beta\}$, (b) initial samples $\{\hx_i\}_{i=1}^{\tilde{n}}$, (c) Langevin samples $\{\tx_i\}_{i=1}^{\tilde{n}}$

\item[]
\State Let $t\leftarrow 0$, randomly initialize $\theta$, $\alpha$, and $\beta$.
\Repeat 
\State {\bf ancestral Langevin sampling}: For $i = 1, ..., \tn$, sample $\hat{z}_i \sim \mathcal{N}(0, I_d)$, then generate $\hat{x}_i = g(\hat{z}_i)$, and run $l$ steps of Langevin revision starting from $\hat{x}_i$ to obtain $\tilde{x}_i$,  each step 
following Eq.~(\ref{eq:Langevin})(ii). 
\State {\bf modified contrastive divergence}: Treat $\{ \tilde{x} \}_i^{\tilde{n}}$ as MCMC examples from $p_{\theta}(x)$, and update $\theta$ by Adam with the gradient computed according to Eq.~(\ref{eq:gradient2}). 
\State {\bf variational MCMC teaching}: Treat  $\{ \tilde{x} \}_i^{\tilde{n}}$ as training data, update $\alpha$ and $\beta$ by minimizing VAE objective in Eq.~(\ref{eq:vae_lower_bound}) via Adam.
\State Let $t \leftarrow t+1$
\Until $t = T$
\end{algorithmic}
\end{algorithm}

\vspace{0.07in}

\begin{figure}[b]
    \centering
    \includegraphics[width=1\linewidth]{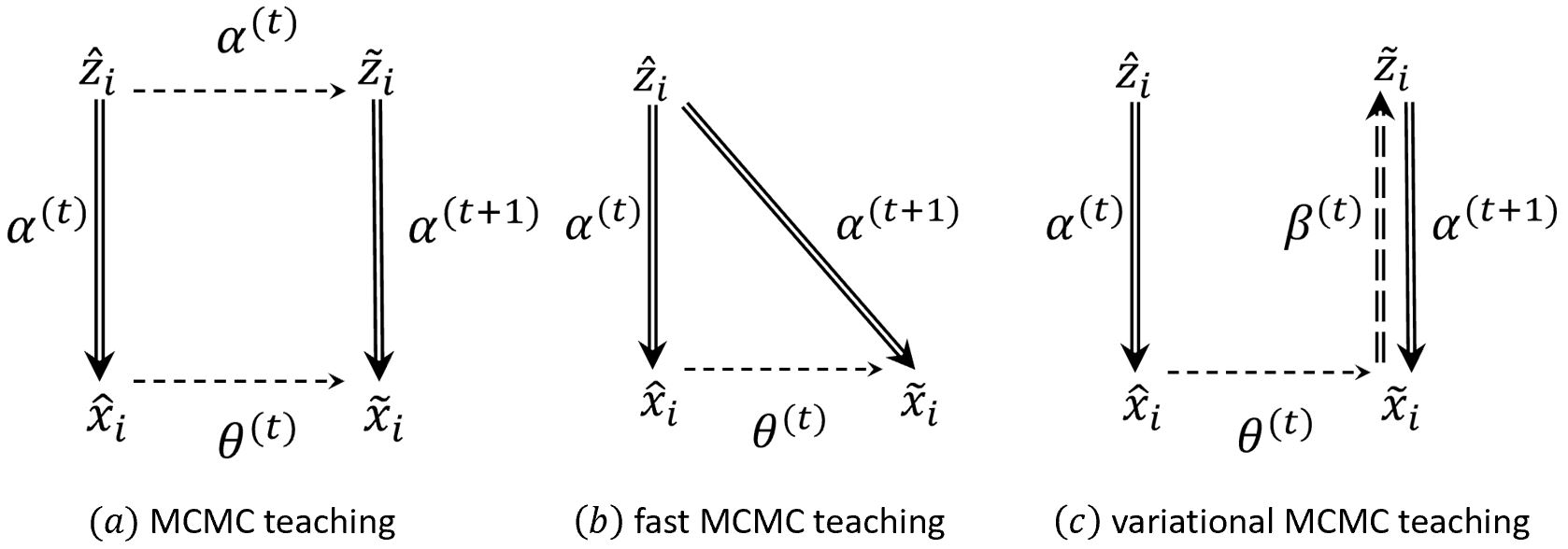}
    \caption{Diagrams of different types of MCMC teaching algorithms. (a) original MCMC teaching with an MCMC-based inference process. (b) fast MCMC teaching without an inference step. (c) \textit{variational MCMC teaching}. The double-solid-line arrows indicate generation and reconstruction by the latent variable model with parameters $\alpha$. The
dashed-line arrows indicate Langevin dynamics guided by $\theta$ in the latent space or data space. The double-dashed-line arrow indicates inference and encoding by the inference model with $\beta$.}
\label{fig:MCMC_teaching}
\end{figure}

 Figure~\ref{fig:MCMC_teaching} shows a comparison of the basic ideas of different types of MCMC teaching strategies. Figure~\ref{fig:MCMC_teaching}(a) and (b) illustrate the diagrams of the original MCMC teaching and its fast variant in~\citet{xie2016cooperative}, respectively. Figure~\ref{fig:MCMC_teaching}(c) displays the proposed \textit{variational MCMC teaching} algorithm. Our framework in Figure~\ref{fig:MCMC_teaching}(c) involves three models and adopts the reparameterization trick for inference, which is different from Figure~\ref{fig:MCMC_teaching}(a) and (b).

\subsection{Optimality of the Solution}
In this section, we present a theoretical understanding of the framework presented in Section~\ref{sec:model}.
A Nash equilibrium of the model is a triplet $(\hat{\theta}, \hat{\alpha}, \hat{\beta})$ that satisfies:
\begin{eqnarray}\label{eq:convergence1}
\resizebox{0.93\hsize}{!}{%
 $\hat{\theta} = \arg \min_{\theta} [ \KL(p_{\rm data}(x){\parallel}p_{\theta}(x)) - \KL(\M_{\hat{\theta}} q_{\hat{\alpha}}(x){\parallel} p_{\theta}(x)) ],$%
 }%
 \end{eqnarray}
\begin{eqnarray} \label{eq:convergence2}
\resizebox{0.93\hsize}{!}{%
$\hat{\alpha}= \arg \min_{\alpha} [ \KL(\M_{\hat{\theta}} q_{\hat{\alpha}}(x){\parallel}q_{\alpha}(x)) +\KL(\pi_{\hat{\beta}}(z|x){\parallel}q_{\alpha}(z|x)) ],$}
\end{eqnarray}
\begin{eqnarray}\label{eq:convergence3}
\hat{\beta}= \arg \min_{\beta}  \KL(\pi_{\beta}(z|x){\parallel}q_{\hat{\alpha}}(z|x)).
\end{eqnarray}

We show that below if $(\hat{\theta}, \hat{\alpha}, \hat{\beta})$ is a Nash equilibrium of the model, then $p_{\hat{\theta}} = q_{\hat{\alpha}} = p_{\rm data}$. 

In Eq.~(\ref{eq:convergence2}) and Eq.~(\ref{eq:convergence3}), the tractable encoder $\pi_{\hat{\beta}}(z|x)$ seeks to approximate the analytically intractable posterior distribution $q_{\hat{\alpha}}(z|x)$ via a joint minimization. When $\KL(\pi_{\hat{\beta}}(z|x){\parallel}q_{\hat{\alpha}}(z|x))=0$, then the second KL-divergence term in Eq.~(\ref{eq:convergence2}) vanishes, thus reducing Eq.~(\ref{eq:convergence2}) to $\min_{\alpha} \KL(\M_{\hat{\theta}} q_{\hat{\alpha}}(x){\parallel}q_{\alpha}(x))$, which means that $q_{\hat{\alpha}}$ seeks to be a stationary distribution of $\M_{\hat{\theta}}$, which is $p_{\hat{\theta}}$. Formally speaking, when $\min_{\alpha}\KL(\M_{\hat{\theta}} q_{\hat{\alpha}}(x)||q_{\alpha}(x))=0$, then $\M_{\hat{\theta}} q_{\hat{\alpha}}(x)=q_{\hat{\alpha}}(x)$, that is,  $q_{\hat{\alpha}}$ converges to the stationary distribution $p_{\hat{\theta}}$, therefore we have $q_{\hat{\alpha}}(x)=p_{\hat{\theta}}(x)$. As a result, the second KL-divergence in Eq.~(\ref{eq:convergence1}) vanishes because $\KL(\M_{\hat{\theta}} q_{\hat{\alpha}}(x){\parallel} p_{\hat{\theta}}(x))=\KL(\M_{\hat{\theta}} p_{\hat{\theta}}(x){\parallel} p_{\hat{\theta}}(x))=0$. 
Eq.~(\ref{eq:convergence1}) is eventually reduced to minimizing the first KL-divergence $\KL(p_{\rm data}(x){\parallel}p_{\hat{\theta}}(x))$, thus, $p_{\hat{\theta}}(x)=p_{\rm data}(x)$. The overall effect of the algorithm is that the \ac{EBM} $p_{\theta}$ runs toward the data distribution $p_{\rm data}$ while inducing the latent variable model $q_{\rm \alpha}$ to get close to the data distribution $p_{\rm data}$ as well, because $q_{\alpha}$ chases $p_{\theta}$ toward $p_{\rm data}$, i.e., $q_{\alpha} \rightarrow p_{\theta} \rightarrow p_{\rm data}$, thus $q_{\hat{\alpha}} = p_{\hat{\theta}} = p_{\rm data}$. In other words, the joint training algorithm can lead to MLE of $q_{\alpha}$ and $p_{\theta}$.

\subsection{Conditional Predictive Learning}
The proposed framework can be generalized to supervised learning of the conditional distribution of an output $x$ given an input $y$, where both input and output are high-dimensional structured variables and may belong to two different modalities. 
We generalize the framework by turning both \ac{EBM} and latent variable model into conditional ones. Specifically, the conditional \ac{EBM} $p_{\theta}(x|y)$ represents a conditional distribution of $x$ given $y$ by using a joint energy function $U_{\theta}(x,y)$, the conditional latent variable model $q_{\alpha}(x|y,z)$ generates $x$ by mapping $y$ and a vector of latent  Gaussian noise variables $z$ together via $x=g_{\alpha}(y,z)$, and the conditional inference network $\pi_{\beta}(z|y,x) \sim \mathcal{N}(\mu_{\beta}(x,y), v_{\beta}(x,y))$, where $\mu_{\beta}(x,y)$ and $v_{\beta}(x,y)$ are outputs of an encoder network taking $x$ and $y$ as inputs.  $q_{\alpha}(x|y,z)$ and $\pi_{\beta}(z|x,y)$ form a conditional \ac{VAE}~\cite{sohn2015learning}. Both the latent variable $z$ in the latent variable model and the Langevin dynamics in the \ac{EBM} allow for randomness in such a conditional mapping, thus making the proposed model suitable for representing one-to-many mapping. Once the conditional model is trained, we can generate samples $\{\tilde{x}_i\}$ conditioned on an input $y$ by following the \textit{ancestral Langevin sampling} process. To use the model on prediction tasks, we can perform a deterministic generation as prediction without sampling, i.e., the conditional latent variable model first generates an initial prediction via $z^*=\E(z), \hat{x}_i=g_{\alpha}(y_i, z^*)$, and then the conditional \ac{EBM} refines $\hat{x}$ by a finite steps of noise-disable Langevin dynamics  $\tilde{x}_{t+1}=\tilde{x}_{t}-\frac{\delta^2}{2} \frac{\partial U(\tilde{x}_t, y_i)}{\partial \tilde{x}}$ with $\tilde{x}_{t=0}=\hat{x}$, which actually is  a gradient descent that finds a local minimum around $\hat{x}$ in the learned energy function $U_{\theta}(x,y=y_i)$.

\section{Information Geometric Understanding}
In this section, we shall provide an information geometric understanding of the proposed learning algorithm, and show that our learning algorithm can be interpreted as a process of dynamic alternating projection within the framework of information geometry. 

\subsection{Three Families of Joint Distributions}
The proposed framework includes three trainable models, i.e., energy-based model $p_{\theta}(x)$, inference model $\pi_{\beta}(z|x)$, and latent variable model $q_{\alpha}(x|z)$. They, along with 
the empirical data distribution $p_{\rm data}(x)$ and the Gaussian prior distribution $q(z)$, define three families of joint distributions over the latent variables $z$ and the data $x$. Let us define
\begin{itemize}
\item $\Pi$-distribution: $\Pi(z,x) =  p_{\rm data}(x) \pi_{\beta}(z|x)$

\item Q-distribution: $Q(z,x) =  q(z) q_{\alpha}(x|z)$

\item P-distribution: $P(z,x) =  p_{\theta}(x) \pi_{\beta}(z|x)$
\end{itemize}
In the context of information geometry, the above three families of distributions can be represented by three different manifolds. Each point of the manifold stands for a probability distribution with a certain parameter.  

The \textit{variational MCMC teaching} that we proposed in this paper to train both EBM and VAE actually integrates variational learning and energy-based learning, which is a modification of maximum
likelihood estimation. The training process alternates these two learning processes, and eventually leads to maximum likelihood solutions of all the models. We first try to understand each part separately below, and then we integrate them together to give a final interpretation. 

\subsection{Variational Learning as Alternating Projection}

\begin{figure}[t]
\centering
\includegraphics[width=.48\linewidth]{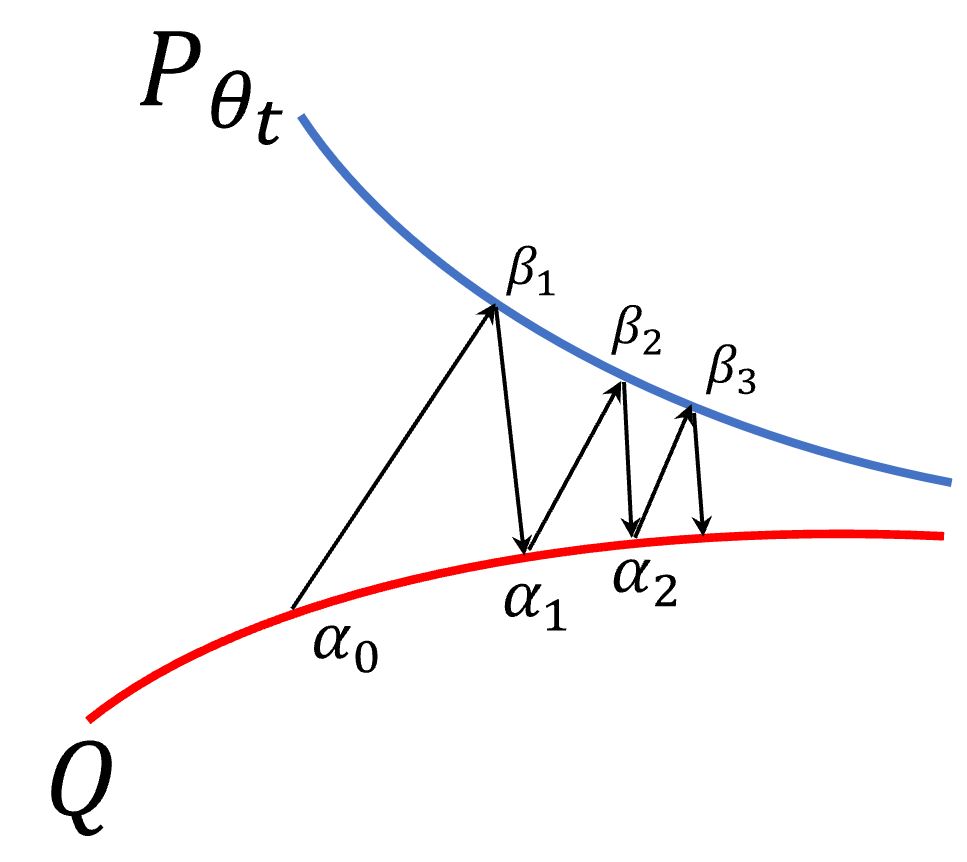}
\caption{Variational learning is interpreted as an process of alternating projection between manifolds $P_{\theta_{t}}$ and $Q$. Manifold $P_{\theta_{t}}$ is represented by a blue curve and manifold $Q$ is represented by a red curve. Each point of the red curve corresponds to a certain $\alpha$, while each point of the blue curve corresponds to a certain $\beta$.}
\label{fig:projection}
\end{figure}

The original variational learning algorithm, such as VAEs~\cite{kingma2013auto}, is to learn $\{\alpha,\beta\}$ from training data $p_{\rm data}(x)$, whose objective function is a joint minimization $\min_{\beta} \min_{\alpha} \KL(\Pi||Q)$. However, in our learning algorithm, the VAE component learns to mimic the EBM at each iteration by learning from its generated examples. Thus, given $\theta_t$ at iteration $t$, our VAE objective becomes $\min_{\beta} \min_{\alpha} \KL(P_{\theta_{t}}||Q)$, where we put subscript $\theta_{t}$ in $P$ to indicate that the P-distribution is associated with a fixed $\theta_{t}$. The following reveals that  $\KL(P_{\theta_{t}}||Q)$ is exactly the VAE loss we use in Eq. (\ref{eq:vae_lower_bound2}). 
\begin{eqnarray}
\begin{aligned}
&\KL(P_{\theta_{t}}||Q)\\
=&\KL(p_{\theta_{t}}(x)\pi_{\beta}(z|x)||q_{\alpha}(x|z)q(z))\\
=&\KL(p_{\theta_{t}}(x)||q_{\alpha}(x)) + \KL(\pi_{\beta}(z|x)||q_{\alpha}(z|x)) \\
=&\KL(\M_{\theta_{t}}q_{\alpha_t}(x)||q_{\alpha}(x)) + \KL(\pi_{\beta}(z|x)||q_{\alpha}(z|x)). \label{eq:vae2}\\
\end{aligned}
\end{eqnarray}
Minimizing the KL-divergence between two probability distributions can be interpreted as a projection from a probability distribution to a manifold \cite{cover1999elements}. 
Therefore, as illustrated in Figure \ref{fig:projection}, each manifold is visualized as a curve and the joint minimization in VAE in Eq.~(\ref{eq:vae2}) can be interpreted as alternating projection \cite{han2019divergence} between manifolds
$P_{\theta_{t}}$ and $Q$, where $\pi_{\beta}$ and $q_{\alpha}$ run toward each other and eventually converge at the intersection between manifolds $P_{\theta_{t}}$~and~$Q$. 

\subsection{Energy-Based Learning as Manifold Shifting}

\begin{figure}[tbh!]
\centering
\includegraphics[width=.48\linewidth]{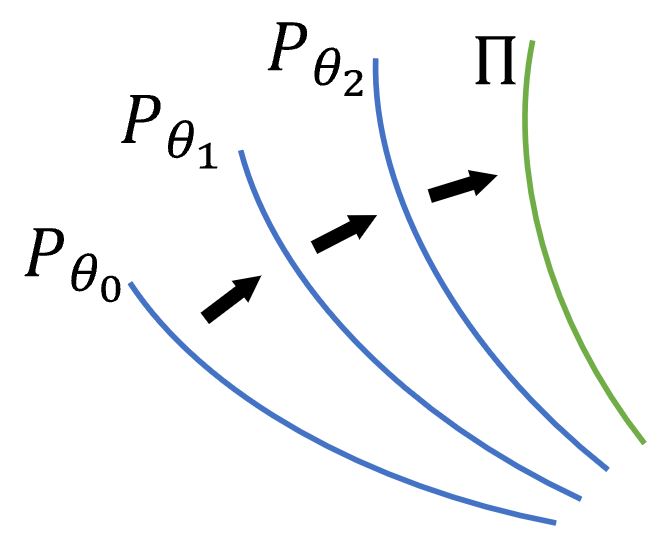}
\caption{Energy-based learning is interpreted as a manifold shifting process from $P_{\theta_0}$ to $\Pi$, where $\theta_0$ denotes the initial $\theta$ at time 0. Manifolds $\{P_{\theta_t}\}$ are represented by blue curves, while manifold $\Pi$ is represented by a green curve.}
\label{fig:shift}
\end{figure}

With the examples generated by the \textit{ancestral Langevin sampler}, the objective function of training the EBM is $\min_{\theta} \KL(\Pi ||P)$, i.e., $\min_{\theta} \KL(p_{\rm data}||p_{\theta})$. As illustrated in Figure~\ref{fig:shift}, $P_{\theta_0}$ runs toward $\Pi$ and seeks to match it. Each point in each curve represents a different $\beta$.

\subsection{Integrating Energy-Based Learning and Variational Learning as Dynamic Alternating Projection}

\vspace{0.08in}

\begin{figure}[h]
\centering
\includegraphics[width=.56\linewidth]{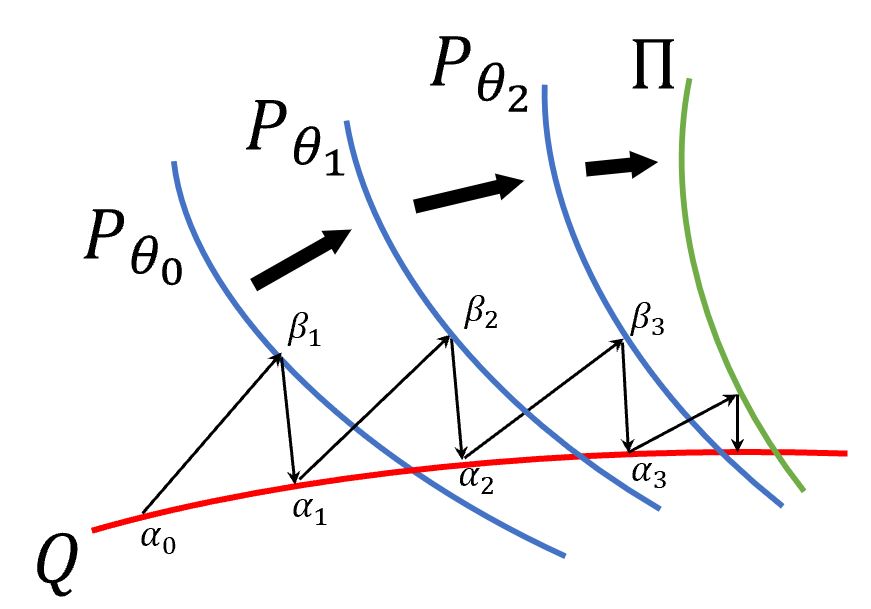}
\caption{\textit{Variational MCMC teaching} as dynamic alternating projection. Manifolds $P$, $Q$, and $\Pi$ are represented by blue, red, and green curves, respectively.}
\label{fig:motional}
\end{figure}

\begin{figure}[h]
\centering
\includegraphics[width=.43\linewidth]{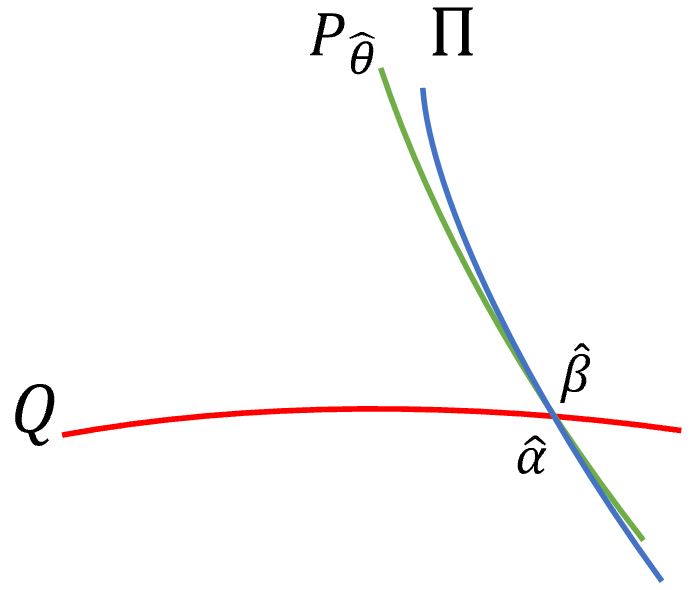}
\caption{Convergent point of the dynamic alternating projection. Triplet ($\hat{\theta}$, $\hat{\alpha}$, $\hat{\beta}$) is the Nash equilibrium (optimal solution) of the learning algorithm. }
\label{fig:converge}
\end{figure}

The joint training of $p_{\theta}$, $\pi_{\beta}$, $q_{\alpha}$ in the proposed framework integrates energy-based learning and variational learning, which can be interpreted as a dynamic alternating projection between $Q$ and $P$, where $Q$ is static but $P$ is changeable and keeps shifting toward $\Pi$. See Figure~\ref{fig:motional} for an illustration. Ideally, $P$ matches $\Pi$, i.e., $P_{\hat{\theta}}=\Pi$. The alternating projection would converge at the intersection point among $Q, P$ and $\Pi$ (see Figure~\ref{fig:converge}), where we have $\min_{\alpha} \min_{\beta} \KL(\Pi||Q)$, which is the objective of the original VAE. In other words, $Q$ and $P$ get close to each other, while $P$ seeks to get close to $\Pi$. In the end, $q_{\alpha}$ chases $p_{\theta}$ towards $p_{\rm data}$.  

\begin{figure*}[h]
\centering
\begin{tabular}{ccc}
\includegraphics[width=.31\linewidth]{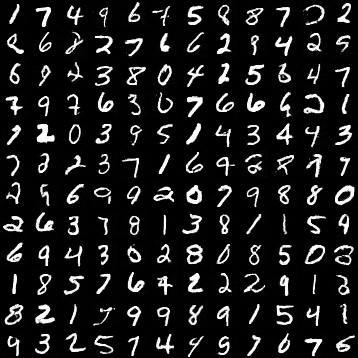} & 
\includegraphics[width=.31\linewidth]{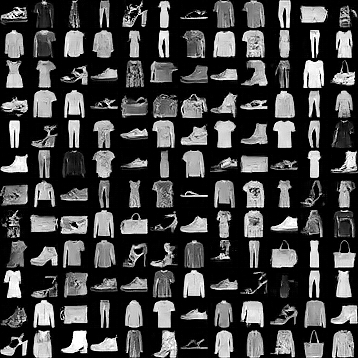} & 
\includegraphics[width=.31\linewidth]{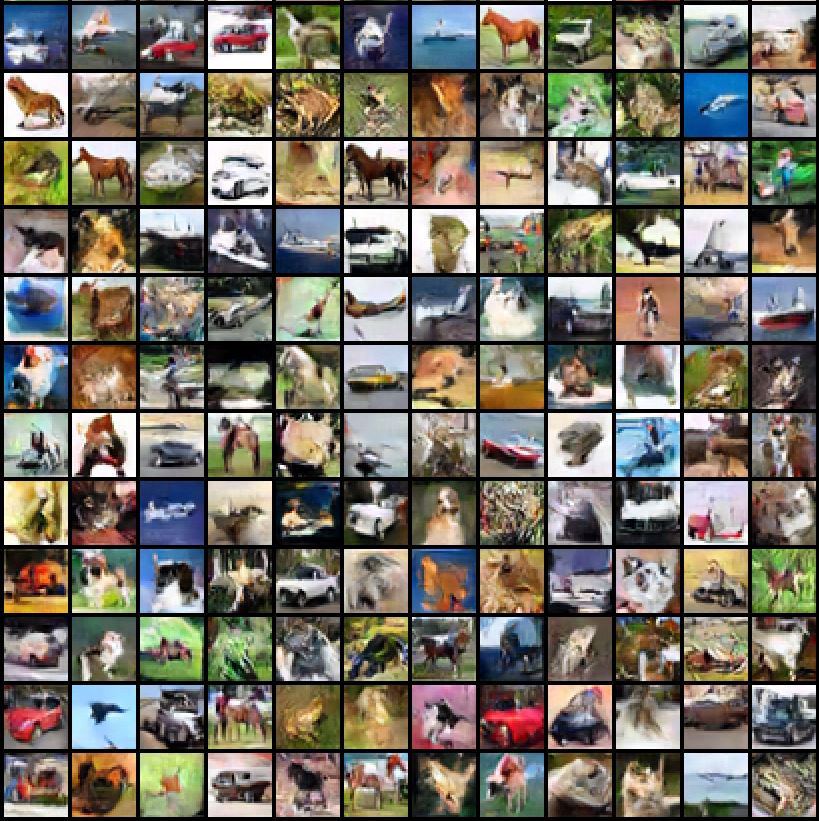} \\
\end{tabular}
\caption{Generated samples by the models learned on MNIST, Fashion-MNIST and CIFAR-10 datasets respectively.}
\label{fig:syn}
\end{figure*}

\subsection{Comparison with Related Models}
We highlight the difference between the proposed method and the closely related models, such as triangle divergence~\cite{han2019divergence} and cooperative network~\cite{xie2018cooperative}. The proposed model optimizes
\begin{eqnarray}\notag
\begin{aligned}
\min_{\theta, \alpha,\beta} \KL(\Pi||P) + \KL(P||Q)
\end{aligned}
\end{eqnarray}
or equivalently 
\[ \min_{\theta, \alpha,\beta} \KL(p_{\rm data}||p_{\theta}) + \KL(p_{\theta}||q_{\alpha}) + \KL(\pi_{\beta}(z|x)||q_{\alpha}(z|x)), \nonumber
\]
which is different from the triangle divergence~\cite{han2019divergence} framework which also trains energy-based model, inference model and latent variable model together but optimizes the following different objective  
\begin{eqnarray}
\begin{aligned}
\min_{\theta, \alpha,\beta} \KL(\Pi||Q) + \KL(Q||P) - \KL(\Pi||P). \nonumber
\end{aligned}
\end{eqnarray}

The cooperative learning~\cite{xie2018cooperative} framework (CoopNets) jointly trains the energy-based model $p_{\theta}(x)$ and the latent variable model $q_{\alpha}(x)$ by 
\[ \min_{\theta, \alpha} \KL(p_{\rm data}||p_{\theta}) + \KL(p_{\theta}||q_{\alpha}),
\]
without leaning an approximate $\pi_{\beta}(z|x)$. Instead, CoopNets~\citep{xie2018cooperative} directly accesses the inference process $q_{\alpha}(z|x)$ by MCMC sampling.

\section{Experiments}
We present experiments to demonstrate  the effectiveness of our strategy to train \ac{EBM}s with (a) competitive synthesis for images, (b) high expressiveness of the learned latent variable model, and (c) strong performance in image completion. We use the PaddlePaddle \footnote{{https://www.paddlepaddle.org.cn}} deep learning platform.

\subsection{Image Generation}
We show that our framework is effective to represent a probability density of images. We demonstrate the learned model can generate realistic image patterns. We learn our model from MNIST \cite{lecun1998gradient}, Fashion-MNIST \cite{xiao2017fashion} and CIFAR-10~\cite{krizhevsky2009learning} images without class labels. Figure~\ref{fig:syn} shows some examples generated by the \textit{ancestral Langevin sampling}. We also  quantitatively evaluate the qualities of the generated images via FID score~\cite{heusel2017gans} and Inception score~\cite{salimans2016improved} in Table~\ref{tab:comp_mnist} and Table~\ref{tab:synthesis}. 
The experiments validate the effectiveness of our model. We design all networks in our model with simple convolution and ReLU layers, and only use 15 or 50 Langevin steps. The Langevin step size $\delta=0.002$. The number of latent dimension $d=200$.


\begin{table}[ht!]
    \centering
   
    \begin{tabular}{lc}
    \hline
         Model & FID  \\ \hline
         GLO~\cite{bojanowski2018optimizing} & 49.60 \\
         CGlow~\cite{liu2019conditional} & 29.64  \\
         CAGlow~\cite{liu2019conditional} & 26.34 \\
         VAE~\cite{kingma2013auto} & 21.85  \\
         DDGM~\cite{Bengio2016} & 30.87  \\
         BEGAN~\cite{berthelot2017began} & 13.54  \\
         EBGAN~\cite{zhao2017energy} & 11.10  \\
         Triangle~\cite{han2019divergence} & 6.77  \\ 
         CoopNets~\cite{xie2018cooperative} & 9.70  \\
         Ours & 8.95 \\
    \hline
    \end{tabular}
     \caption{Comparison with baseline models on MNIST dataset with respect to FID score ($l=50$).}
    \label{tab:comp_mnist}
\end{table}

\begin{table}[ht!]
\centering
\begin{tabular}{lc}

\hline
Model    & IS    \\ \hline 
PixelCNN~\cite{van2016conditional} & 4.60       \\ 
PixelIQN~\cite{ostrovski2018autoregressive} & 5.29       \\
EBM~\cite{du2019implicit}    & 6.02       \\ 
DCGAN~\cite{radford2015unsupervised}   & 6.40      \\ 
WGAN+GP~\cite{gulrajani2017improved}  & 6.50        \\ 
CoopNets~\cite{xie2016cooperative}   &           6.55      \\ 
Ours     &    6.65        \\
\hline
\end{tabular}
\caption{Inception scores on CIFAR-10 dataset ($l=15$). }
\label{tab:synthesis}
\end{table}

\begin{figure}[h]
\center
\includegraphics[width=1\linewidth]{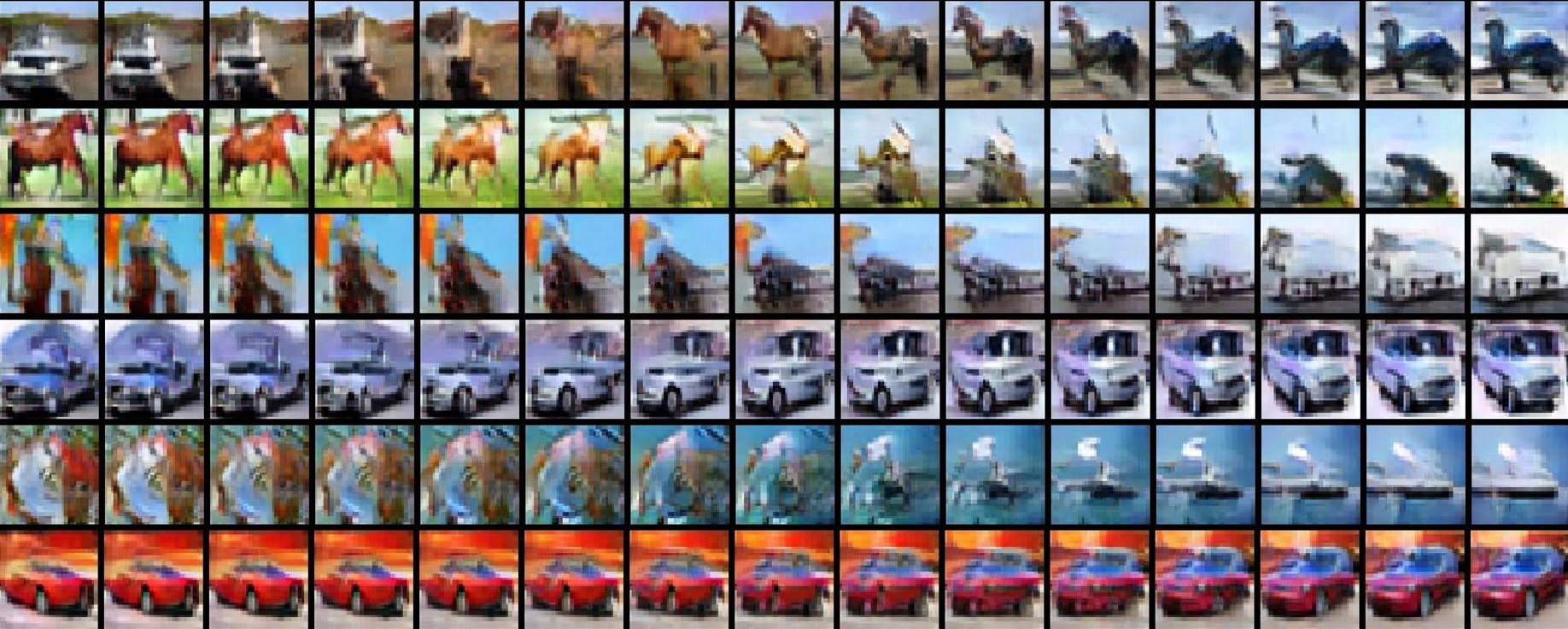} \\
(a) Interpolation by the latent variable model
\includegraphics[width=1\linewidth]{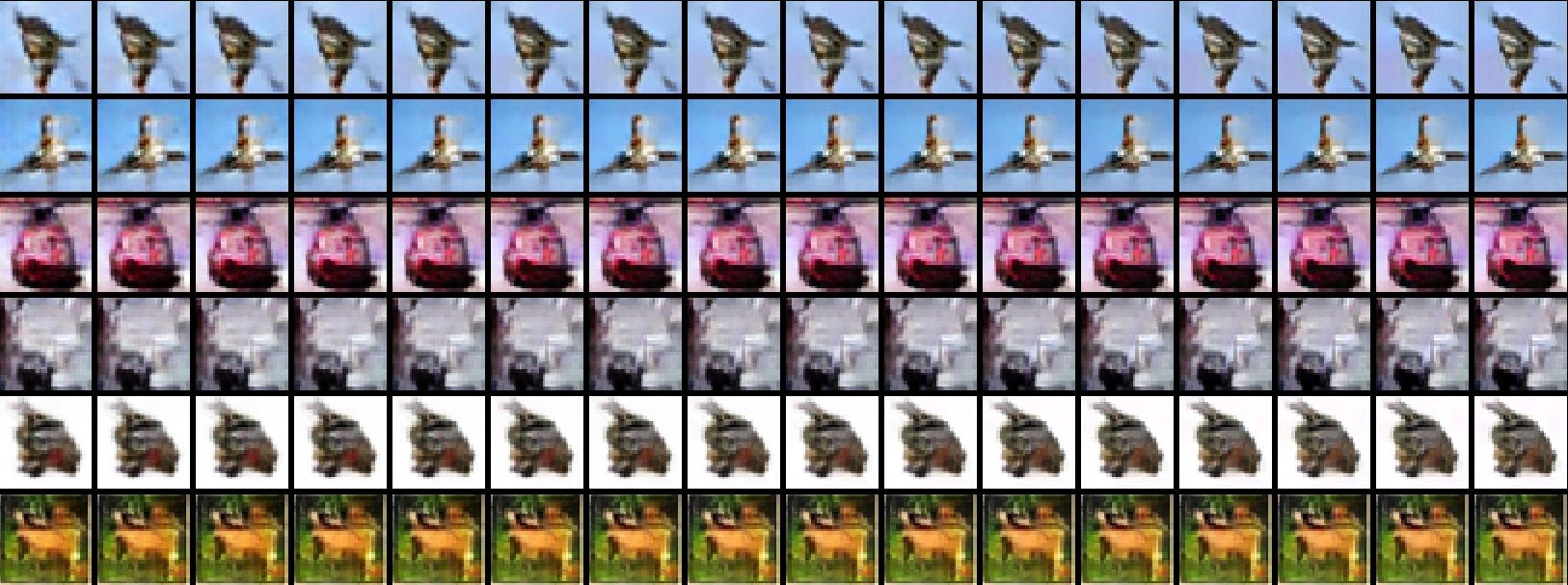} \\
(b) Langevin revision by the learned model 
\caption{Model analysis. (a) Interpolation between latent vectors of the images
on the two ends. (b) Visualization of ancestral Langevin dynamics when the model converges. 
For each row, the leftmost image is the synthesized output
by the ancestral sampling. The rest image sequence displays the synthesized images revised at different Langevin steps. 
}
\label{fig:model}
\end{figure}

\begin{figure*}[t]
	\centering
	\setlength{\tabcolsep}{1.5pt}
	\setlength{\imagewidth}{.118\linewidth}
    \begin{tabular}{c:cccccc:c}
         Input & pix2pix  & cVAE-GAN  & BicycleGAN  & cCoopNets & cVAE-GAN++ & ours  & ground truth \\
         \includegraphics[width=\imagewidth]{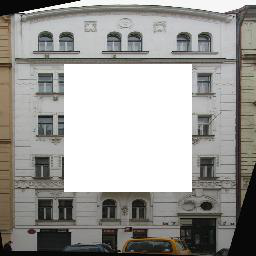} &
         \includegraphics[width=\imagewidth]{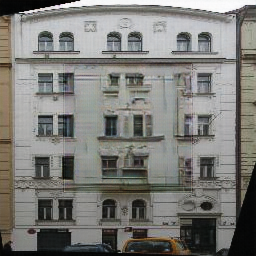} &
         \includegraphics[width=\imagewidth]{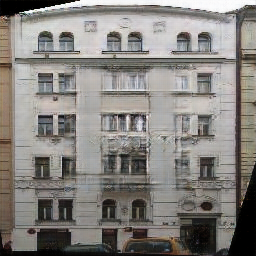} &
         \includegraphics[width=\imagewidth]{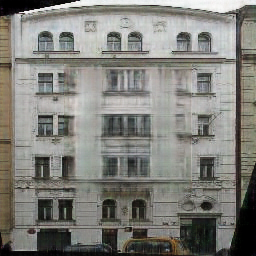} &
         \includegraphics[width=\imagewidth]{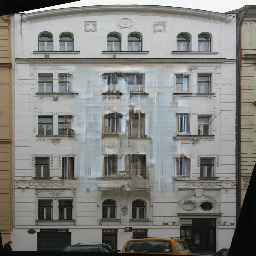} &
         \includegraphics[width=\imagewidth]{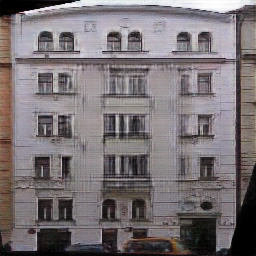} &
         \includegraphics[width=\imagewidth]{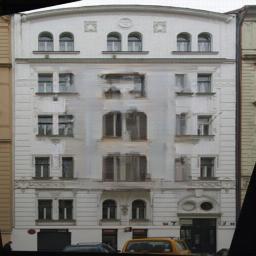} & 
         \includegraphics[width=\imagewidth]{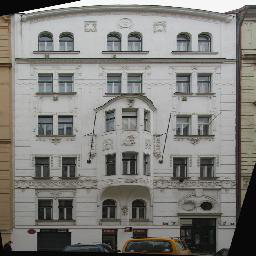} \\
         
         \includegraphics[width=\imagewidth]{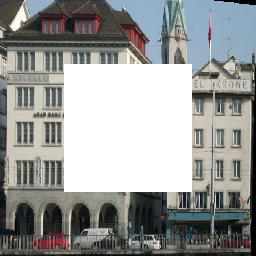} &
         \includegraphics[width=\imagewidth]{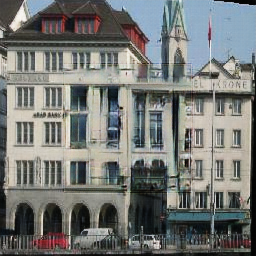} &
         \includegraphics[width=\imagewidth]{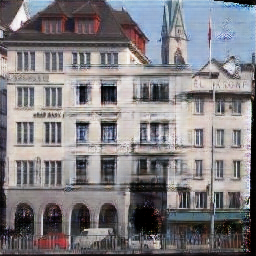} &
         \includegraphics[width=\imagewidth]{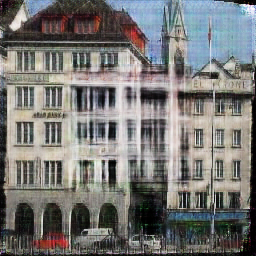} &
         \includegraphics[width=\imagewidth]{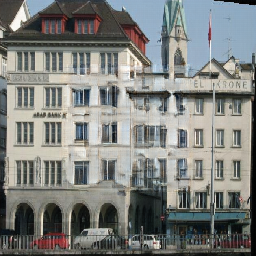} &
         \includegraphics[width=\imagewidth]{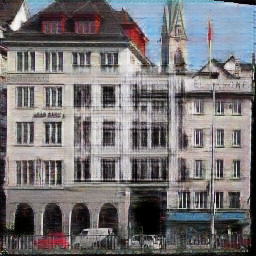} &
         \includegraphics[width=\imagewidth]{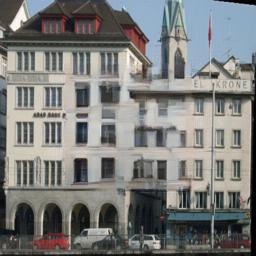} &
         \includegraphics[width=\imagewidth]{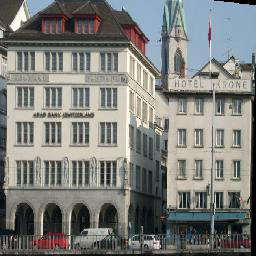} \\
    \end{tabular}
	\caption{Example results of image completion on the Facade test dataset.}
	\label{fig:exp_recovery}
\end{figure*}

We also check whether the latent variable model $q_{\alpha}(x|z)$ learns a meaningful latent space $z$ in this learning scheme by demonstrating interpolation between generated examples in the latent space as shown in Figure~\ref{fig:model}(a). Each row of transition is a sequence of $g_{\alpha}(z_{\eta})$ with interpolated $z_{\eta}= \eta z_l + \sqrt{1-\eta^2} z_r $ where $\eta \in [0,1]$, $z_l$ and $z_r$ are the latent variables of the examples at the left and right ends respectively. The transitions appear smooth, which means that the latent variable model learns a meaningful image embedding. We also check the gap between $p_{\theta}$ and $q_{\alpha}$ once the model is leaned, by visualizing the Langevin dynamics initialized by a sample from the latent variable model in Figure~\ref{fig:model}(b). Each row shows one example, in which the leftmost image is generated by the latent variable model via ancestral sampling, and the rest image sequence shows the revised examples at different Langevin steps. The rightmost one is the final synthesized example after 15 steps of Langevin revision. We can find that even though the Langevin dynamics can still improve the initial example (we can carefully compare the leftmost and the rightmost images, the rightmost one is a little bit sharper than the leftmost one), but their difference is quite small, which is in fact a good phenomenon revealing that the latent variable model has caught up with the \ac{EBM}, which runs toward the data distribution. That is, $q_{\alpha}$ becomes the stationary distribution of $p_{\theta}$, or $\KL(\M_{\hat{\theta}}q_{\hat{\alpha}}(x)||q_{\hat{\alpha}}(x)) \rightarrow 0$.

\subsection{Image Completion}
We apply our conditional model to image completion, where we learn a stochastic mapping from a centrally masked image to the original one. The centrally masked image is of the size $256 \times 256$ pixels, centrally overlaid with a mask of the size $128 \times 128$ pixels. The conditional energy function in $p_{\theta}(x|y)$ takes the concatenation of the masked image $y$ and the original image $x$ as input and consists of three convolutional layers and one fully-connected layer. For the conditional latent variable model $q_{\alpha}(x|y, z)$, we follow~\citet{isola2017image} to use a U-Net~\cite{ronneberger2015u}, with the latent vector $z$ concatenated with its bottleneck. We set $d=8$. The conditional encoder $\pi_{\beta}(z|y,x)$ has five residual blocks and MLP layers to get the variational encoding. We compare our method with baselines including pix2pix~\cite{isola2017image}, cVAE-GAN~\cite{zhu2017toward}, cVAE-GAN++~\cite{zhu2017toward}, BicycleGAN~\cite{zhu2017toward}, and cCoopNets~\cite{xie2016cooperative} on the Paris StreetView~\cite{pathak2016context} and the CMP Facade datasets~\cite{tylevcek2013spatial} in Table~\ref{tab:recovery}. The recovery performance is measured by the peak signal-to-noise ratio (PSNR) and Structural SIMilarity (SSIM) between the recovered image and the original image. Our method outperforms the baselines. Figure~\ref{fig:exp_recovery} shows some qualitative~results.

\begin{table}[ht]
\centering


\begin{tabular}{lcccc}
\hline
\multirow{2}{*}{Method} & \multicolumn{2}{c}{Facade} & \multicolumn{2}{c}{StreetView} \\
 & PSNR & SSIM & PSNR & SSIM \\ \hline 
		pix2pix & 19.34 & 0.74 & 15.17 & 0.75 \\
		cVAE-GAN & 19.43 & 0.68 & 16.12 & 0.72 \\
		cVAE-GAN++ & 19.14 & 0.64 & 16.03& 0.69 \\
		BicycleGAN & 19.07 & 0.64 & 16.00& 0.68 \\
		cCoopNets & 20.47 & 0.77 & 21.17 & 0.79 \\ 
		Ours  & \textbf{21.62} & \textbf{0.78} & \textbf{22.61 }& \textbf{0.79} \\ 
\hline
\end{tabular}
\caption{Comparison with baselines for image completion. }
\label{tab:recovery}
\end{table}


\subsection{Model Analysis}
Our framework involves three different components, each of which has some key hyper-parameters that might affect the behavior of the whole training process. We investigate some factors that may potentially influence the performance of our framework on CIFAR-10. The results are reported after 1,000 epochs of training.

\subsubsection{Number of Langevin steps and Lagevin step size} We first study how the number of Langevin steps and their step size affect the synthesis performance. Table~\ref{tab:ablation_langevin} shows the influence of varying number of Langevin step and Langevin step size, respectively. As the number of Langevin steps increases and the step size decreases, we observe improved quality of image synthesis in terms of inception score.

\begin{table}[h]
\centering
\begin{tabular}{c|ccccc}
\hline 
IS $\uparrow$& $l$ = 5  & $l$ = 8 & $l$ = 15 & $l$ = 30 & $l$ = 60 \\ \hline
$\delta$ = 0.001 & 3.606 &   4.333     &  6.072       &    6.038    & 6.143       \\ 
$\delta$ = 0.002 & 3.847 &   5.568    &  6.075       & 5.989      & 5.882      \\ 
$\delta$ = 0.004 & 4.799 & 5.286      &  5.979       & 5.907       & 5.933       \\ 
$\delta$ = 0.008  & 5.146 &  5.164  &   5.835   & 4.574  & 3.482      \\ 
\hline 
\end{tabular}
\caption{Influence of number of MCMC steps $l$ and MCMC step size~$\delta$, with the number of latent dimension $d=200$, and variational loss penalty $\gamma=2$.}
\label{tab:ablation_langevin}
\end{table}

\subsubsection{Number of dimensions of the latent space} We also study how the number of dimensions of the latent space affect the \textit{ancestral Langevin sampling} process in training the energy-based model. Table~\ref{tab:ablation_latent} displays the inception scores as a function of the number of latent dimensions of $q_{\alpha}{(x)}$. We set $l=10$, $\delta=0.002$, and $\gamma=2$.

\begin{table}[h]
\centering
\begin{tabular}{c|cccccc}
\hline
$d$ & $1200$ & $600$ & $200$ & $100$ & $50$ & $10$ \\ \hline
IS $\uparrow$& 6.017 & 6.213 & 6.159 & 6.085 & 6.027 & 5.973 \\ 
\hline
\end{tabular}
\caption{Influence of the number of latent dimension $d$}
\label{tab:ablation_latent}
\end{table}

\subsubsection{Variational loss penalty} The penalty weight $\gamma$ of the term of KL-divergence between the inference model and the posterior distribution in Eq. (\ref{eq:vae_lower_bound}) plays an important role in adjusting the tradeoff between having low auto-encoding reconstruction loss and having good approximation of the posterior distribution. Table~\ref{tab:ablation_vaeloss} displays the inception scores of varying $\gamma$, with $d=200$, $l=10$, and $\delta=0.002$. The optimal choice of $\gamma$ in our model is roughly 2. 

\begin{table}[h]
\centering
\begin{tabular}{c|cccccccc}
\hline
$\gamma$ & $0.05$ &  $0.5$ & $1$ & $2$ & $8$ & $10$ \\ \hline
IS $\uparrow$ & 5.106 &  5.663 & 5.905 & 6.159 & 5.890 & 4.693 \\ 
\hline
\end{tabular}
\caption{Influence of the variational loss penalty $\gamma$ }
\label{tab:ablation_vaeloss}
\end{table}

\vspace{-1.5mm}

\section{Conclusion}
This paper proposes to learn an \ac{EBM} with a \ac{VAE} as an amortized sampler for probability density estimation. In particular, we propose the \textit{variational MCMC teaching} algorithm to train the EBM and VAE together. In the proposed joint training framework, the latent variable model in the VAE and the Langevin dynamics derived from the EBM learn to collaborate to form an efficient sampler, which is essential to provide Monte Carlo samples to train both the \ac{EBM} and the VAE. The proposed method naturally unifies the maximum likelihood estimation,  variational learning, and MCMC teaching in a single computational framework, and can be interpreted as a dynamic alternating projection within the framework of information geometry. 
Our framework is appealing as it combines the representational flexibility and ability of the \ac{EBM} and the computational tractability and efficiency of the VAE. Experiments show that the proposed framework can be effective in image generation, and its conditional generalization can be useful for computer vision applications, such as image completion.

\bibliography{mybibfile}

\end{document}

%% file: config.tex
\usepackage{amsmath}
\usepackage{algpseudocode}
\usepackage{algorithm}
\usepackage{graphicx}
\usepackage{xcolor}
\usepackage{acronym}
\usepackage{arydshln}
\usepackage{subcaption}
\usepackage{multirow}
\usepackage{xspace}

\def\KL{{\rm KL}}

\def\M{{\cal M}}

\def\hx{\hat{x}}
\def\tx{\tilde{x}}
\def\tn{\tilde{n}}

\def\E{{\rm E}}

\def\tn{\tilde{n}}

\makeatletter
\DeclareRobustCommand\onedot{\futurelet\@let@token\@onedot}
\def\@onedot{\ifx\@let@token.\else.\null\fi\xspace}

\makeatother

\DeclareMathOperator*{\argmin}{arg\,min}

\newlength{\imagewidth}

\makeatletter
\newcommand{\thickhline}{%
    \noalign {\ifnum 0=`}\fi \hrule height 1pt
    \futurelet \reserved@a \@xhline
}
\makeatother

\newacro{EBM}{energy-based model}
\newacro{EBMs}{energy-based models}
\newacro{VAE}{variational auto-encoder}
\newacro{VAEs}{variational auto-encoders}

%% file: camera_ready.bbl
\begin{thebibliography}{57}
\providecommand{\natexlab}[1]{#1}
\providecommand{\url}[1]{\texttt{#1}}
\providecommand{\urlprefix}{URL }
\expandafter\ifx\csname urlstyle\endcsname\relax
  \providecommand{\doi}[1]{doi:\discretionary{}{}{}#1}\else
  \providecommand{\doi}{doi:\discretionary{}{}{}\begingroup
  \urlstyle{rm}\Url}\fi

\bibitem[{Bakhtin et~al.(2021)Bakhtin, Deng, Gross, Ott, Ranzato, and
  Szlam}]{bakhtin2021residual}
Bakhtin, A.; Deng, Y.; Gross, S.; Ott, M.; Ranzato, M.; and Szlam, A. 2021.
\newblock Residual energy-Based models for text.
\newblock \emph{Journal of Machine Learning Research (JMLR)} 22(40): 1--41.

\bibitem[{Barbu and Zhu(2020)}]{barbu2020monte}
Barbu, A.; and Zhu, S.-C. 2020.
\newblock \emph{Monte Carlo methods}.
\newblock Springer.

\bibitem[{Berthelot, Schumm, and Metz(2017)}]{berthelot2017began}
Berthelot, D.; Schumm, T.; and Metz, L. 2017.
\newblock BEGAN: Boundary equilibrium generative adversarial networks.
\newblock \emph{arXiv preprint arXiv:1703.10717} .

\bibitem[{Bojanowski et~al.(2018)Bojanowski, Joulin, Lopez-Pas, and
  Szlam}]{bojanowski2018optimizing}
Bojanowski, P.; Joulin, A.; Lopez-Pas, D.; and Szlam, A. 2018.
\newblock Optimizing the latent space of generative networks.
\newblock In \emph{International Conference on Machine Learning (ICML)},
  600--609.

\bibitem[{Cover(1999)}]{cover1999elements}
Cover, T.~M. 1999.
\newblock \emph{Elements of Information Theory}.
\newblock John Wiley \& Sons.

\bibitem[{Du et~al.(2019)Du, Meier, Ma, Fergus, and Rives}]{du2019energy}
Du, Y.; Meier, J.; Ma, J.; Fergus, R.; and Rives, A. 2019.
\newblock Energy-based models for atomic-resolution protein conformations.
\newblock In \emph{International Conference on Learning Representations
  (ICLR)}.

\bibitem[{Du and Mordatch(2019)}]{du2019implicit}
Du, Y.; and Mordatch, I. 2019.
\newblock Implicit generation and modeling with energy based models.
\newblock In \emph{Advances in Neural Information Processing Systems
  (NeurIPS)}, volume~32, 3608--3618.

\bibitem[{Gao et~al.(2018)Gao, Lu, Zhou, Zhu, and Nian~Wu}]{gao2018learning}
Gao, R.; Lu, Y.; Zhou, J.; Zhu, S.-C.; and Nian~Wu, Y. 2018.
\newblock Learning generative convnets via multi-grid modeling and sampling.
\newblock In \emph{Proceedings of the IEEE Conference on Computer Vision and
  Pattern Recognition (CVPR)}, 9155--9164.

\bibitem[{Geman and Geman(1984)}]{geman1984stochastic}
Geman, S.; and Geman, D. 1984.
\newblock Stochastic relaxation, Gibbs distributions, and the Bayesian
  restoration of images.
\newblock \emph{IEEE Transactions on Pattern Analysis and Machine Intelligence
  (TPAMI)} 6(6): 721--741.

\bibitem[{Grathwohl et~al.(2019)Grathwohl, Wang, Jacobsen, Duvenaud, Norouzi,
  and Swersky}]{grathwohl2019your}
Grathwohl, W.; Wang, K.-C.; Jacobsen, J.-H.; Duvenaud, D.; Norouzi, M.; and
  Swersky, K. 2019.
\newblock Your classifier is secretly an energy based model and you should
  treat it like one.
\newblock In \emph{International Conference on Learning Representations
  (ICLR)}.

\bibitem[{Grenander and Miller(2007)}]{grenander2007pattern}
Grenander, U.; and Miller, M.~I. 2007.
\newblock \emph{Pattern Theory: From Representation to Inference}.
\newblock Oxford University Press.

\bibitem[{Gulrajani et~al.(2017)Gulrajani, Ahmed, Arjovsky, Dumoulin, and
  Courville}]{gulrajani2017improved}
Gulrajani, I.; Ahmed, F.; Arjovsky, M.; Dumoulin, V.; and Courville, A.~C.
  2017.
\newblock Improved training of wasserstein GANs.
\newblock In \emph{Advances in Neural Information Processing Systems (NIPS)},
  5767--5777.

\bibitem[{Gutmann and Hyv{\"a}rinen(2012)}]{gutmann2012noise}
Gutmann, M.~U.; and Hyv{\"a}rinen, A. 2012.
\newblock Noise-contrastive estimation of unnormalized statistical models, with
  applications to natural image statistics.
\newblock \emph{Journal of Machine Learning Research (JMLR)} 13(2): 307--361.

\bibitem[{Han et~al.(2017)Han, Lu, Zhu, and Wu}]{han2017alternating}
Han, T.; Lu, Y.; Zhu, S.-C.; and Wu, Y.~N. 2017.
\newblock Alternating back-propagation for generator network.
\newblock In \emph{Proceedings of the Thirty-First AAAI Conference on
  Artificial Intelligence (AAAI)}, 1976--1984.

\bibitem[{Han et~al.(2019)Han, Nijkamp, Fang, Hill, Zhu, and
  Wu}]{han2019divergence}
Han, T.; Nijkamp, E.; Fang, X.; Hill, M.; Zhu, S.-C.; and Wu, Y.~N. 2019.
\newblock Divergence triangle for joint training of generator model,
  energy-based model, and inferential model.
\newblock In \emph{Proceedings of the IEEE Conference on Computer Vision and
  Pattern Recognition (CVPR)}, 8670--8679.

\bibitem[{Heusel et~al.(2017)Heusel, Ramsauer, Unterthiner, Nessler, and
  Hochreiter}]{heusel2017gans}
Heusel, M.; Ramsauer, H.; Unterthiner, T.; Nessler, B.; and Hochreiter, S.
  2017.
\newblock GANs trained by a two time-scale update rule converge to a local Nash
  equilibrium.
\newblock In \emph{Advances in Neural Information Processing Systems (NIPS)},
  6626--6637.

\bibitem[{Hinton(2002)}]{Hinton2002a}
Hinton, G.~E. 2002.
\newblock {Training products of experts by minimizing contrastive divergence.}
\newblock \emph{Neural Computation} 14(8): 1771--1800.

\bibitem[{Hinton(2012)}]{hinton2012practical}
Hinton, G.~E. 2012.
\newblock A practical guide to training restricted Boltzmann machines.
\newblock In \emph{Neural Networks: Tricks of the Trade}, 599--619.

\bibitem[{Ingraham et~al.(2018)Ingraham, Riesselman, Sander, and
  Marks}]{ingraham2018learning}
Ingraham, J.; Riesselman, A.; Sander, C.; and Marks, D. 2018.
\newblock Learning protein structure with a differentiable simulator.
\newblock In \emph{International Conference on Learning Representations
  (ICLR)}.

\bibitem[{Isola et~al.(2017)Isola, Zhu, Zhou, and Efros}]{isola2017image}
Isola, P.; Zhu, J.-Y.; Zhou, T.; and Efros, A.~A. 2017.
\newblock Image-to-image translation with conditional adversarial networks.
\newblock In \emph{Proceedings of the IEEE Conference on Computer Vision and
  Pattern Recognition (CVPR)}, 1125--1134.

\bibitem[{Kim and Bengio(2016)}]{Bengio2016}
Kim, T.; and Bengio, Y. 2016.
\newblock Deep directed generative models with energy-based probability
  estimation.
\newblock \emph{arXiv preprint arXiv:1606.03439} .

\bibitem[{Kingma and Ba(2015)}]{kingma2014adam}
Kingma, D.~P.; and Ba, J. 2015.
\newblock Adam: A method for stochastic optimization.
\newblock \emph{International Conference on Learning Representations (ICLR)} .

\bibitem[{Kingma and Welling(2014)}]{kingma2013auto}
Kingma, D.~P.; and Welling, M. 2014.
\newblock Auto-encoding variational Bayes.
\newblock In \emph{International Conference on Learning Representations
  (ICLR)}.

\bibitem[{Krizhevsky(2009)}]{krizhevsky2009learning}
Krizhevsky, A. 2009.
\newblock Learning multiple layers of features from tiny images.
\newblock Technical report, University of Toronto.

\bibitem[{LeCun et~al.(1998)LeCun, Bottou, Bengio, and
  Haffner}]{lecun1998gradient}
LeCun, Y.; Bottou, L.; Bengio, Y.; and Haffner, P. 1998.
\newblock Gradient-based learning applied to document recognition.
\newblock \emph{Proceedings of the IEEE} 86(11): 2278--2324.

\bibitem[{LeCun et~al.(2006)LeCun, Chopra, Hadsell, Ranzato, and
  Huang}]{Lecun2006}
LeCun, Y.; Chopra, S.; Hadsell, R.; Ranzato, M.; and Huang, F.~J. 2006.
\newblock A tutorial on energy-based learning.
\newblock In \emph{Predicting Structured Data}. MIT Press.

\bibitem[{Liu(2008)}]{liu2008monte}
Liu, J.~S. 2008.
\newblock \emph{Monte Carlo strategies in scientific computing}.
\newblock Springer Science \& Business Media.

\bibitem[{Liu et~al.(2019)Liu, Liu, Gong, Wang, and Li}]{liu2019conditional}
Liu, R.; Liu, Y.; Gong, X.; Wang, X.; and Li, H. 2019.
\newblock Conditional adversarial generative flow for controllable image
  synthesis.
\newblock In \emph{Proceedings of the IEEE Conference on Computer Vision and
  Pattern Recognition (CVPR)}, 7992--8001.

\bibitem[{Lu, Zhu, and Wu(2016)}]{LuZhuWu2016}
Lu, Y.; Zhu, S.-C.; and Wu, Y.~N. 2016.
\newblock Learning FRAME models using CNN filters.
\newblock In \emph{Thirtieth AAAI Conference on Artificial Intelligence
  (AAAI)}, 1902--1910.

\bibitem[{Neal(2011)}]{neal2011mcmc}
Neal, R.~M. 2011.
\newblock MCMC using Hamiltonian dynamics.
\newblock \emph{Handbook of Markov Chain Monte Carlo} 2.

\bibitem[{Nijkamp et~al.(2019)Nijkamp, Hill, Zhu, and Wu}]{nijkamp2019learning}
Nijkamp, E.; Hill, M.; Zhu, S.-C.; and Wu, Y.~N. 2019.
\newblock Learning non-convergent non-persistent short-run MCMC toward
  energy-based model.
\newblock In \emph{Advances in Neural Information Processing Systems
  (NeurIPS)}, 5233--5243.

\bibitem[{Ostrovski, Dabney, and Munos(2018)}]{ostrovski2018autoregressive}
Ostrovski, G.; Dabney, W.; and Munos, R. 2018.
\newblock Autoregressive quantile networks for generative modeling.
\newblock In \emph{International Conference on Machine Learning (ICML)},
  3936--3945.

\bibitem[{Pathak et~al.(2016)Pathak, Krahenbuhl, Donahue, Darrell, and
  Efros}]{pathak2016context}
Pathak, D.; Krahenbuhl, P.; Donahue, J.; Darrell, T.; and Efros, A.~A. 2016.
\newblock Context encoders: Feature learning by inpainting.
\newblock In \emph{Proceedings of the IEEE Conference on Computer Vision and
  Pattern Recognition (CVPR)}, 2536--2544.

\bibitem[{Radford, Metz, and Chintala(2016)}]{radford2015unsupervised}
Radford, A.; Metz, L.; and Chintala, S. 2016.
\newblock Unsupervised representation learning with deep convolutional
  generative adversarial networks.
\newblock \emph{International Conference on Learning Representations (ICLR)} .

\bibitem[{Ronneberger, Fischer, and Brox(2015)}]{ronneberger2015u}
Ronneberger, O.; Fischer, P.; and Brox, T. 2015.
\newblock U-net: Convolutional networks for biomedical image segmentation.
\newblock In \emph{International Conference on Medical Image Computing and
  Computer Assisted Intervention}, 234--241.

\bibitem[{Salimans et~al.(2016)Salimans, Goodfellow, Zaremba, Cheung, Radford,
  and Chen}]{salimans2016improved}
Salimans, T.; Goodfellow, I.; Zaremba, W.; Cheung, V.; Radford, A.; and Chen,
  X. 2016.
\newblock Improved techniques for training GANs.
\newblock In \emph{Advances in Neural Information Processing Systems (NIPS)},
  2226--2234.

\bibitem[{Sohn, Lee, and Yan(2015)}]{sohn2015learning}
Sohn, K.; Lee, H.; and Yan, X. 2015.
\newblock Learning structured output representation using deep conditional
  generative models.
\newblock In \emph{Advances in Neural Information Processing Systems (NIPS)},
  3483--3491.

\bibitem[{Song and Ou(2018)}]{song2018learning}
Song, Y.; and Ou, Z. 2018.
\newblock Learning neural random fields with inclusive auxiliary generators.
\newblock \emph{arXiv preprint arXiv:1806.00271} .

\bibitem[{Tyle{\v{c}}ek and {\v{S}}{\'a}ra(2013)}]{tylevcek2013spatial}
Tyle{\v{c}}ek, R.; and {\v{S}}{\'a}ra, R. 2013.
\newblock Spatial pattern templates for recognition of objects with regular
  structure.
\newblock In \emph{German Conference on Pattern Recognition (GCPR)}, 364--374.

\bibitem[{Van~den Oord et~al.(2016)Van~den Oord, Kalchbrenner, Espeholt,
  Vinyals, and Graves}]{van2016conditional}
Van~den Oord, A.; Kalchbrenner, N.; Espeholt, L.; Vinyals, O.; and Graves, A.
  2016.
\newblock Conditional image generation with pixelCNN decoders.
\newblock In \emph{Advances in Neural Information Processing Systems (NIPS)},
  4790--4798.

\bibitem[{Wu, Zhu, and Liu(2000)}]{wu2000texture}
Wu, Y.~N.; Zhu, S.-C.; and Liu, X. 2000.
\newblock Equivalence of Julesz ensembles and FRAME models.
\newblock \emph{International Journal of Computer Vision (IJCV)} 38: 247--265.

\bibitem[{Xiao, Rasul, and Vollgraf(2017)}]{xiao2017fashion}
Xiao, H.; Rasul, K.; and Vollgraf, R. 2017.
\newblock Fashion-MNIST: A novel image dataset for benchmarking machine
  learning algorithms.
\newblock \emph{arXiv preprint arXiv:1708.07747} .

\bibitem[{Xie et~al.(2014)Xie, Hu, Zhu, and Wu}]{xie2014learning}
Xie, J.; Hu, W.; Zhu, S.-C.; and Wu, Y.~N. 2014.
\newblock Learning sparse FRAME models for natural image patterns.
\newblock \emph{International Journal of Computer Vision (IJCV)} 1--22.

\bibitem[{Xie et~al.(2018{\natexlab{a}})Xie, Lu, Gao, and
  Wu}]{xie2016cooperative}
Xie, J.; Lu, Y.; Gao, R.; and Wu, Y.~N. 2018{\natexlab{a}}.
\newblock Cooperative learning of energy-based model and latent variable model
  via MCMC teaching.
\newblock In \emph{Proceedings of the AAAI Conference on Artificial
  Intelligence (AAAI)}, 4292--4301.

\bibitem[{Xie et~al.(2018{\natexlab{b}})Xie, Lu, Gao, Zhu, and
  Wu}]{xie2018cooperative}
Xie, J.; Lu, Y.; Gao, R.; Zhu, S.-C.; and Wu, Y.~N. 2018{\natexlab{b}}.
\newblock Cooperative training of descriptor and generator networks.
\newblock \emph{IEEE Transactions on Pattern Analysis and Machine Intelligence
  (TPAMI)} 42(1): 27--45.

\bibitem[{Xie et~al.(2016)Xie, Lu, Zhu, and Wu}]{XieLuICML}
Xie, J.; Lu, Y.; Zhu, S.-C.; and Wu, Y. 2016.
\newblock A theory of generative ConvNet.
\newblock In \emph{International Conference on Machine Learning (ICML)},
  2635--2644.

\bibitem[{Xie et~al.(2021{\natexlab{a}})Xie, Xu, Zheng, Zhu, and
  Wu}]{xie2021GPointNet}
Xie, J.; Xu, Y.; Zheng, Z.; Zhu, S.; and Wu, Y.~N. 2021{\natexlab{a}}.
\newblock Generative PointNet: energy-based learning on unordered point sets
  for 3D generation, reconstruction and classification.
\newblock In \emph{Proceedings of the IEEE Conference on Computer Vision and
  Pattern Recognition (CVPR)}.

\bibitem[{Xie et~al.(2019)Xie, Zheng, Fang, Zhu, and Wu}]{xie2019cooperative}
Xie, J.; Zheng, Z.; Fang, X.; Zhu, S.-C.; and Wu, Y.~N. 2019.
\newblock Cooperative training of fast thinking initializer and slow thinking
  solver for multi-modal conditional learning.
\newblock \emph{arXiv preprint arXiv:1902.02812} .

\bibitem[{Xie et~al.(2021{\natexlab{b}})Xie, Zheng, Fang, Zhu, and
  Wu}]{xie2021cycleCoopNets}
Xie, J.; Zheng, Z.; Fang, X.; Zhu, S.-C.; and Wu, Y.~N. 2021{\natexlab{b}}.
\newblock Learning cycle-consistent cooperative networks via alternating MCMC
  teaching for unsupervised cross-domain translation.
\newblock In \emph{Proceedings of The Thirty-Fifth AAAI Conference on
  Artificial Intelligence (AAAI)}.

\bibitem[{Xie et~al.(2018{\natexlab{c}})Xie, Zheng, Gao, Wang, Zhu, and
  Wu}]{xie2018learning}
Xie, J.; Zheng, Z.; Gao, R.; Wang, W.; Zhu, S.-C.; and Wu, Y.~N.
  2018{\natexlab{c}}.
\newblock Learning descriptor networks for 3D shape synthesis and analysis.
\newblock In \emph{Proceedings of the IEEE Conference on Computer Vision and
  Pattern Recognition (CVPR)}, 8629--8638.

\bibitem[{Xie et~al.(2020)Xie, Zheng, Gao, Wang, Zhu, and
  Wu}]{Xie2020GenerativeVL}
Xie, J.; Zheng, Z.; Gao, R.; Wang, W.; Zhu, S.-C.; and Wu, Y.~N. 2020.
\newblock Generative VoxelNet: learning energy-based models for 3D shape
  synthesis and analysis.
\newblock \emph{IEEE Transactions on Pattern Analysis and Machine Intelligence
  (TPAMI)} .

\bibitem[{Xie, Zhu, and Wu(2017)}]{XieCVPR17}
Xie, J.; Zhu, S.-C.; and Wu, Y.~N. 2017.
\newblock Synthesizing dynamic patterns by spatial-temporal generative ConvNet.
\newblock In \emph{Proceedings of the IEEE Conference on Computer Vision and
  Pattern Recognition (CVPR)}, 7093--7101.

\bibitem[{Xie, Zhu, and Wu(2019)}]{xie2019learning}
Xie, J.; Zhu, S.-C.; and Wu, Y.~N. 2019.
\newblock Learning energy-based spatial-temporal generative convnets for
  dynamic patterns.
\newblock \emph{IEEE Transactions on Pattern Analysis and Machine Intelligence
  (TPAMI)} .

\bibitem[{Xu et~al.(2019)Xu, Xie, Zhao, Baker, Zhao, and Wu}]{xu2019energy}
Xu, Y.; Xie, J.; Zhao, T.; Baker, C.; Zhao, Y.; and Wu, Y.~N. 2019.
\newblock Energy-based continuous inverse optimal control.
\newblock \emph{arXiv preprint arXiv:1904.05453} .

\bibitem[{Zhao, Mathieu, and LeCun(2017)}]{zhao2017energy}
Zhao, J.; Mathieu, M.; and LeCun, Y. 2017.
\newblock Energy-based generative adversarial networks.
\newblock In \emph{International Conference on Learning Representations
  (ICLR)}.

\bibitem[{Zhu et~al.(2017)Zhu, Zhang, Pathak, Darrell, Efros, Wang, and
  Shechtman}]{zhu2017toward}
Zhu, J.-Y.; Zhang, R.; Pathak, D.; Darrell, T.; Efros, A.~A.; Wang, O.; and
  Shechtman, E. 2017.
\newblock Toward multimodal image-to-image translation.
\newblock In \emph{Advances in Neural Information Processing Systems (NIPS)},
  465--476.

\bibitem[{Zhu, Wu, and Mumford(1998)}]{zhu1998filters}
Zhu, S.~C.; Wu, Y.; and Mumford, D. 1998.
\newblock Filters, random fields and maximum entropy {(FRAME)}: Towards a
  unified theory for texture modeling.
\newblock \emph{International Journal of Computer Vision (IJCV)} 27(2):
  107--126.

\end{thebibliography}
